\documentclass[letterpaper, 10 pt, conference]{ieeeconf}  % Comment this line out if you need a4paper

\IEEEoverridecommandlockouts                              % This command is only needed if 
\usepackage{xcolor}
\usepackage{graphicx}
\usepackage[misc]{ifsym}
\usepackage{amsmath,amsfonts}
\usepackage{algorithmic}
\usepackage{array}
\usepackage[compatibility=false]{caption}\usepackage{subcaption}
\usepackage{textcomp}
\usepackage{stfloats}
\usepackage{url}
\usepackage{verbatim}
\usepackage{booktabs}
\usepackage{graphicx}
\usepackage{siunitx}
\usepackage{hyperref}
\usepackage{multirow}
\usepackage{comment}
\usepackage{graphicx}% Include figure files
\usepackage{dcolumn}% Align table columns on decimal point
\usepackage{bm}% bold math
\usepackage{ulem}

\title{\LARGE \bf
Bio-Inspired Pneumatic Modular Actuator for Peristaltic Transport}

\author{Brian Ye$^{\dag}$, Zhuonan Hao$^{\dag}$, Priya Shah, Mohammad Khalid Jawed$^{*}$ % <-this % stops a space
\thanks{$\dag$ The authors are equal contributions.}% <-this % stops a space
\thanks{$*$ Corresponding author. Email: \tt khalidjm@seas.ucla.edu}
\thanks{All authors are with Department of Mechanical and Aerospace Engineering, University of California, Los Angeles, 420 Westwood Plaza, Los Angeles, CA 90095.} %
}

\begin{document}

\maketitle
\thispagestyle{empty}
\pagestyle{empty}

%%%%%%%%%%%%%%%%%%%%%%%%%%%%%%%%%%%%%%%%%%%%%%%%%%%%%%%%%%%%%%%%%%%%%%%%%%%%%%%%
\begin{abstract}

%%%
While its biological significance is well-documented, its application in soft robotics, particularly for the transport of fragile and irregularly shaped objects, remains underexplored. 
%%%
This study presents a modular soft robotic actuator system that addresses these challenges through a scalable, adaptable, and repairable framework, offering a cost-effective solution for versatile applications.
%%%
The system integrates optimized donut-shaped actuation modules and utilizes real-time pressure feedback for synchronized operation, ensuring efficient object grasping and transport without relying on intricate sensing or control algorithms. 
%%%
Experimental results validate the system’s ability to accommodate objects with varying geometries and material characteristics, balancing robustness with flexibility. 
%%%
This work advances the principles of peristaltic actuation, establishing a pathway for safely and reliably manipulating delicate materials in a range of scenarios.
\end{abstract}

\noindent \textbf{Keywords:} Bio-inspired system, peristaltic transport, soft robotics, pneumatic actuator

%%%%%%%%%%%%%%%%%%%%%%%%%%%%%%%%%%%%%%%%%%%%%%%%%%%%%%%%%%%%%%%%%%%%%%%%%%%%%%%%
\section{Introduction}
Peristalsis, defined as the involuntary, wave-like contraction and relaxation of circular and longitudinal muscles~\cite{patel2023physiology}, is a widespread biological mechanism essential for various functions in animals and humans.
%%%
It serves two primary purposes: generating locomotion and facilitating the transport of substances (e.g., food and fluids). 
%%%
On one hand, soft-bodied animals, such as earthworms, sea cucumbers, snails, and certain types of fish, use peristalsis for locomotion~\cite{quillin1999kinematic, nakamura2008locomotion}. 
%%%
They rely on wave-like muscular elongation and contraction along their bodies to propel themselves through their environment~\cite{tanaka2012mechanics}, such as soil, water, or along surfaces. 
%%%
On the other hand, peristalsis is fundamental for biological organisms to transport solid or liquid substances~\cite{bursian2012structure}, facilitated by the propagation of rhythmic contraction waves along tubular structures~\cite{brasseur1987fluid}, such as the esophagus, stomach, intestines, and ureter~\cite{gora2016tubular}.
%%%
The process provides slow but stable~\cite{nakamura2008locomotion} and adaptable transportation~\cite{saga2004development}, minimizing energy consumption and enabling movement through small or irregular openings~\cite{seok2012meshworm} while maintaining functionality under varying conditions~\cite{sensoy2021review}.

\begin{figure}[!ht]
\centering
\includegraphics[width=0.5\textwidth]{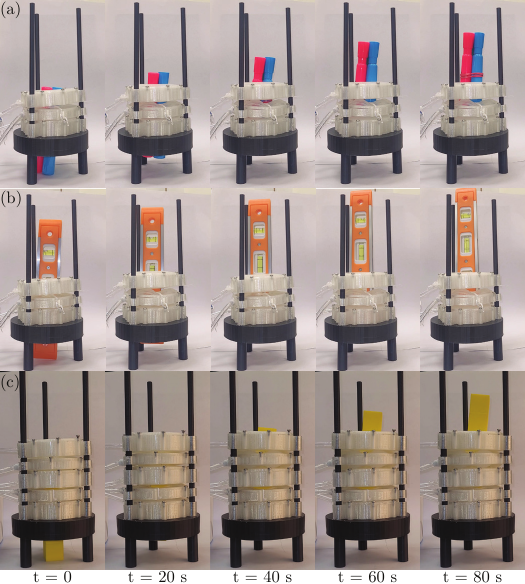}
\caption{Overview of the actuator's capability to grasp delicate and irregular objects, demonstrated using a fundamental unit (three stacks) for: (a) a bundle of marker pens, (b) a handheld tool (leveler), and a multi-level unit (five stacks) for (c) a fragile soft cuboid.}
\label{fig:overview}
\end{figure}

With recent advancements in soft robotics, researchers have increasingly focused on developing soft robots and bio-inspired systems that replicate biological mechanisms and behaviors~\cite{trivedi2008soft, rus2015design}.
%%%
To leverage the benefits of a soft, compliant body and the peristalsis mechanism, worm-like crawling robots have been introduced as effective locomotion generators~\cite{seok2012meshworm, nakamura2008locomotion, mangan2002development, menciassi2004sma, saga2004development, nakamura2006development, nakamura2006development1, saga2004prototype, kim2005earthworm}. 
%%%
These studies have addressed various aspects including actuator design (e.g., SMA, servo motors, pneumatics, magnetic fields), fabrication methods (e.g., silicone polymers, fiber coils, electroactive polymers, artificial muscles), and control strategies (e.g., thermal feedback, pressure feedback, potentiometer feedback). 
%%%
However, the application of the peristalsis mechanism for transporting objects through artificial tubular structures remains underexplored.
%%% 
Relevant works in this area primarily focus on monolithic device design~\cite{peng2024peristaltic, dirven2013design}, particularly in applications such as swallowing or digestive systems. 
%%%
However, these systems lack adaptability when object size varies, such as in the case of elongated or irregularly shaped objects. 
%%%
Furthermore, when a part of the system malfunctions, the entire system becomes faulty, requiring time and material to replace or reproduce the integrated system, which can be a significant drawback in terms of efficiency and reliability. 
%%%
This limitation underscores the need for more adaptable and modular systems that allow for localized repairs or replacements without disrupting the entire functionality.
%%%
Other studies~\cite{hashem2020design, dang2019soft} demonstrate the use of donut-shaped actuators that function as grippers capable of handling delicate, soft food items, such as cherry tomatoes and soft candies, without causing damage.
%%%
However, these studies do not investigate the use of peristalsis for transporting objects.
%%%
Therefore, our objective is to address the challenges of grasping and transporting fragile and irregularly shaped objects, highlighting the potential of soft robotic systems for safely handling sensitive materials.
%%%
As the \autoref{fig:overview} shown, such a system must be modular, allowing for length adjustments tailored to specific tasks, easy to transport and assemble for rapid deployment, and resilient so that the failure of a single module does not compromise the entire operation.

\begin{figure*} [ht]
    \centering
    \includegraphics[width=1\textwidth]{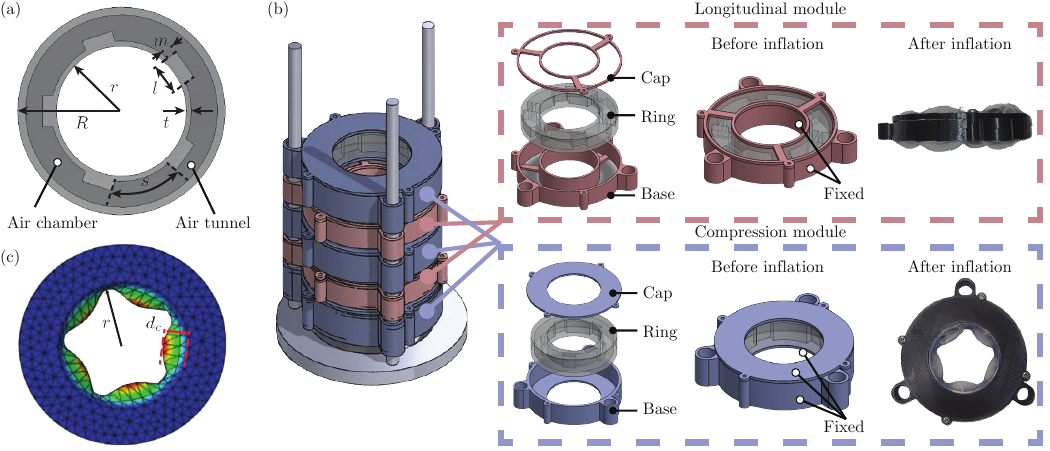}
    \caption{Design of the actuation module. (a) The actuation ring (cross-sectional view) is donut-shaped with outer radius ($R$), inner radius ($r$), and multiple air chambers characterized by length ($s$), spacing ($l$), and wall thickness ($t$). Chambers are connected by air tunnels with step height ($m$). (b) The actuator station consists of a stack of different actuation modules. In longitudinal module, casing allows inflation along the top and bottom surfaces. For compression module, casing only allows inflation on the inner surface. (c) Finite element mesh of the ring.}
    \label{fig:design}
\end{figure*}

In this paper, we present a bio-inspired, pneumatic, modular peristaltic actuator system. 
%%%
Our contributions include:
%%%
\begin{enumerate}
    \item The geometrical optimization of a single donut-shaped actuation module, along with the development of a modular actuator system that can be easily assembled, adjusted in length, and repaired in situ.
    \item A real-time pressure feedback control strategy to ensure efficient object handling and transportation.
\end{enumerate}

The paper outlines the design, fabrication, and control of the peristaltic actuator system.
%%%
We provide a thorough analysis of the actuator's performance, emphasizing the synchronization of the actuation modules through real-time pressure feedback.
%%%
The paper is structured as follows: Section~\ref{sec:PeristalticActuator} provides an overview of the actuator design and material specifications, followed by a detailed discussion on the manufacturing and assembly process.
%%%
Section~\ref{sec:peristalticLocomotion}, we delve into the peristaltic locomotion mechanism and its operational strategies.
%%%
Finally, Section~\ref{sec:Conclusions} presents the experimental and simulation results, followed by a discussion of the key findings.

\section{Peristaltic Actuator}
\label{sec:PeristalticActuator}

This section details the design and fabrication of the proposed bio-inspired peristaltic actuator. The actuator features two distinct types of inflation modules: a compression module, which produces radial displacement, and a longitudinal module, which generates axial displacement. The modular design enables easy stacking of multiple units in any configuration or sequence, allowing for enhanced scalability and adaptability.

% The following subsections detail the design, fabrication, modeling, and assembly, providing an overview of the actuator's architecture.

\subsection{Design and Material Specifications}

Each actuation module is composed of donut-shaped soft rings containing multiple internal air chambers, encased in a 3D-printed outer casing, as shown in \autoref{fig:design}. To simplify fabrication, both types of actuation modules -- compression and longitudinal -- utilize a similar ring design as shown in \autoref{fig:design}(b). The strain-limiting casing restricts the deformation of the soft ring in specific directions: the compression module is prevented from axial deformation, while the longitudinal module is constrained from radial deformation. As a result, the compression module only inflates radially to grasp an object, while the longitudinal module inflates axially to transport it along the axis of the stacked modules, hereafter referred to as the ``station." This modular approach allows for low-cost replacements in the event of partial module failures, eliminating the need for multiple specialized molds.

The inclusion of multiple air chambers and tunnels in \autoref{fig:design}(a) is essential for promoting uniform inflation of the ring. Using a single air chamber can lead to visible inhomogeneous deformation, resulting in a loss of axisymmetry due to inherent experimental and fabrication errors, as the rings are never completely uniform in thickness. By incorporating multiple air chambers, we aim to prevent defects in one chamber from adversely affecting the inflation of all the other chambers.

The rings are fabricated from ``Ecoflex 00-45 Near Clear," a rubber-like silicone material selected for its low cost and hyperelastic behavior with relatively low stiffness (Shore 00 hardness 45) that facilitates easy inflation. Its hyperelastic nature permits significant deformations over numerous cycles without permanent shape change, enabling effective pneumatic actuation and ensuring consistent performance without degradation over time. The outer casing is constructed from Thermoplastic Polyurethane (TPU), which has a higher stiffness (Shore A hardness 95). TPU serves as a strain-limiting layer, providing the necessary structural support to withstand the weight of stacked modules in the station. Without TPU, the stacked soft rings would be unable to maintain the station\rq{}s overall shape due to their own weight. 

\begin{figure*}[ht]
    \centering
    \includegraphics[width=1\linewidth]{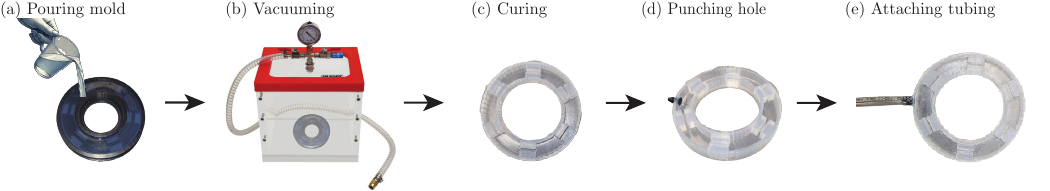}
    \caption{Fabrication process of the soft actuation ring. (a) Filling the mold with silicone material. (b) Eliminating air bubbles using a vacuum. (c) Solidifying the silicone. (d) Creating openings for connectors and tubing on the outer surface. (e) Connecting tubing to complete the assembly. (f) Repeating steps a and b with the upper mold, curing together with the completed bottom mold. }
    \label{fig:fabrication}
\end{figure*}

\subsection{Manufacturing and Assembly Process}

The base and cap of the TPU casings, as shown in \autoref{fig:design}(b), are printed using a 3D printer (Original Prusa i3 MK3S+ 3D Printer). Following this, we fabricate the soft rings through a step-by-step casting and curing process, as illustrated in \autoref{fig:fabrication}. To create the mixture, Part A and Part B of Ecoflex are combined in a 1:1 ratio by weight or volume. To fabricate the internal air chambers, the ring mold is designed with separate top and bottom sections. Once the two sections are glued together later in the process, the empty spaces in each section will combine to form the air chambers. 

First, the mixture is poured into the bottom section to form the bottom surface and walls. To eliminate air bubbles from the mixture, the entire assembly (mixture and mold) is placed in a vacuum chamber for two minutes. As the mixture begins to solidify after a few minutes, a small hole is punctured in the outer surface to create an opening for the tubing connector used for inflation. The tubing connector is then attached, and additional Ecoflex mixture is used to seal the connection between the tubing and the bottom section. After approximately 20 minutes, the mixture starts to take on a solid form. At this stage, but before it fully cures (which can take several hours), the bottom part is placed onto the base of the TPU casing, as depicted in \autoref{fig:design}(b).

Additional Ecoflex mixture is prepared and poured into the mold for the top section. Vacuum chamber is again used to remove air bubbles. The partially cured bottom section is then positioned directly onto the still-curing top section. We allow this assembly to cure for an hour or more, ensuring a robust bond between the two parts and proper sealing of the air chambers. By allowing the top section to cure simultaneously with the bottom, we strengthen the mechanical bond between the two surfaces, enhancing the overall durability and performance of the actuator module.

The final stage involves connecting the TPU cap shown in \autoref{fig:design}(b) to the base using mechanical screws. Once each actuation module is completed, we stack them in the sequence of compression, longitudinal, compression, and so on, to construct an integrated actuator station. To ensure alignment among multiple modules, 3D-printed guide rails are used.

\subsection{Simulation Framework}

The inflation behavior of the actuator was simulated using Finite Element Analysis (FEA) in Abaqus\footnote{\url{https://github.com/StructuresComp/Peristaltic_Actuator.git}}. The geometry was discretized using tetrahedral mesh elements (C3D10), with a 3D model created in SolidWorks and imported into Abaqus. Representative geometric parameters for the experimental setup are detailed in \autoref{tab:para}. The mesh included 49,736 nodes and 29,686 elements. A hyperelastic Mooney-Rivlin model was used, with nonlinear geometry enabled in the software. Material properties were approximated for "Ecoflex 00-45 Near Clear" (Shore 00 hardness 45), assigning a Young's modulus of 100 $\si{kPa}$~\cite{kim2011epidermal} and a Poisson's ratio of 0.45 (selected to avoid numerical instability from a Poisson ratio of 0.5 while remaining close to incompressible behavior). The model's coefficients were $C_{10} = C_{01} =$ 8620.69 \si{Pa} (assumed equal) and $D_1 = 2.4 \times 10^{-6}$ \si{Pa}. A pressure load of 15 \si{kPa}, measured experimentally, was applied to simulate the inflation process.

\begin{table}[h!]
\centering
\caption{Geometrical parameters}
\label{tab:para}
\begin{tabular}{cccc}
\toprule
Description & Symbol & Value & Unit \\
\midrule
Outer radius & $R$ & $40$ & \si{mm} \\ 
Inner radius & $r$ & $25$ & \si{mm} \\ 
Step height & $m$ &  $4$ & \si{mm} \\ 
Chamber spacing & $l$ & $12$ & \si{mm} \\ 
Wall thickness & $t$ & $2$ & \si{mm} \\ 
Chamber length & $s$ &  $28.8$ & \si{mm} \\ 
\bottomrule
\end{tabular}
\end{table}

The boundary conditions were applied differently for the two modules. For the longitudinal module, as shown in \autoref{fig:design}(c), nodes on the side walls were constrained using an encastre boundary condition, meaning they were fully fixed in place with no translation or rotation allowed, to prevent any deformation. The nodes on the top and bottom walls were free to deform under the applied pressure. In contrast, for the compression module, the nodes on the top and bottom walls, as well as the nodes on the outer surface of the side walls, were constrained. However, the nodes on the inner side walls were allowed to deform.

For the purposes of numerical simulations, keep in mind that the primary function of the longitudinal module is to provide upward movement through repeated actuation cycles. The total displacement of the object is simply the product of the upward movement per cycle times the number of cycles. Since this is a prototype design study, our primary focus is not on power consumption and efficiency at this stage but rather on feasibility. We can simply increase the number of cycles to achieve the desired movement. In contrast, the compression module is crucial for grasping the object securely during transport. A stronger grasp allows the system to handle a wider variety of objects. Therefore, the focus of our modeling efforts is on refining the geometric parameters of the compression module to maximize its grasping effectiveness. To describe the inflation behavior, we define a parameter called normalized inflation as the ratio of the maximum inflation height $d_c$ to the inner radius of the ring $r$, as \autoref{fig:design}(c). The parameter $d_c$ can be interpreted as the maximum radial displacement of the nodes located on the inner side wall.

\subsection{System-Level Hardware Setup}

\begin{figure} [ht]
    \centering
    \includegraphics[width=1\linewidth]{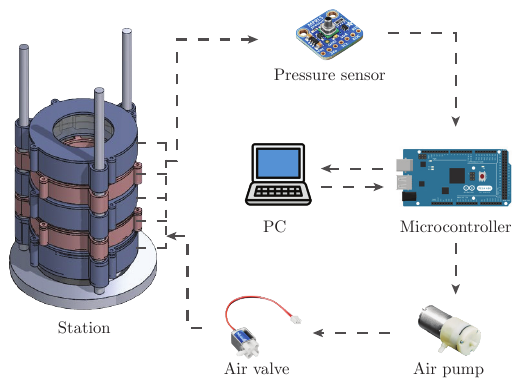}
    \caption{Experimental setup. The actuator station integrates air pumps, air valves, pressure sensors, a microcontroller, and a computer, creating a closed-loop control system for actuator operation and data acquisition.}
    \label{fig:setup}
\end{figure}

The overall setup relies on sequential inflation and deflation cycles of the modules to transport an object, as detailed later in Section~\ref{sec:peristalticLocomotion}. This process requires precise, closed-loop pressure control for each module, allowing the system to detect when the object is in contact with a module through variations in pressure signals. As shown in \autoref{fig:setup}, the system to accomplish this objective includes pressure sensors, a microcontroller, a computer, air pumps, and air valves. This forms a closed-loop system for sensing, control, and data acquisition. 

The pressure sensors (Adafruit MPRLS Pressure Sensor) monitor the internal pressures of each actuation module in real time, providing data for feedback loops to control inflation and deflation cycles. This pressure data is transmitted to the microcontroller (Arduino Mega 2560 Rev3), which manages airflow supplied by the air pump (Adafruit Air Pump and Vacuum DC Motor -- ZR370-02PM).

\begin{figure*}[t]
\centering
\includegraphics[width=\linewidth]{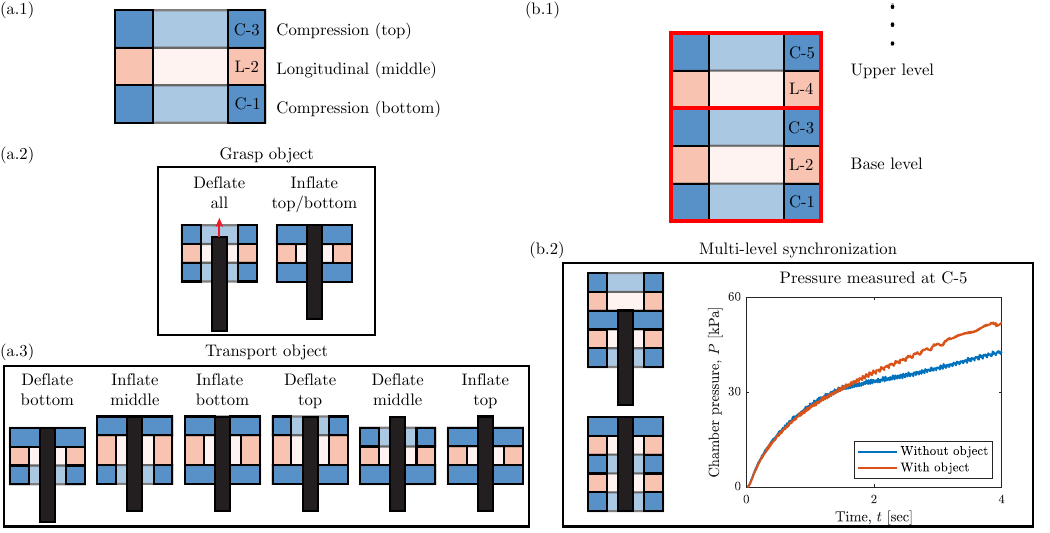}
\caption{Control of peristaltic locomotion. (a.1) Fundamental control module consisting of two compression modules positioned at the top and bottom, with a longitudinal module in the middle. (a.2) Control sequence to grasp object. (a.3) Control sequence to transport object. (b.1) The multi-level actuator station. (b.2) Multi-level synchronization based on pressure signal differences between the upper level, comparing scenarios with and without object grasp.}
\label{fig:control}
\end{figure*}

Following a control sequence outlined later in \autoref{fig:control}, the air valves (Adafruit 6V Air Valve) precisely regulate airflow into and out of the individual actuation rings. The pressure data is then collected and analyzed on a connected computer.

\section{Peristaltic Locomotion}
\label{sec:peristalticLocomotion}

The peristaltic locomotion mechanism in our actuator leverages sequential inflation and deflation cycles of the individual actuation modules to achieve controlled movement and object manipulation. This section introduces two key aspects of the system's operation. First, the fundamental working unit, composed of three actuation modules, is used to perform basic tasks like object grasp and transport through sequential control. Second, we explore the concept of synchronization as additional modules are stacked together, forming a multi-level structure that requires coordinated control across multiple layers. This ensures consistent and efficient behavior, facilitated by a closed-loop control system.

\subsection{Three-Module Actuation Sequence}

The fundamental working unit for peristaltic locomotion consists of three actuation modules: a compression module at the bottom (C--1), a longitudinal module in the middle (L--2), and another compression module at the top (C--3), as \autoref{fig:control}(a.1). This configuration allows for sequential inflation and deflation to achieve object grasp and transport.

To begin, both the top and bottom compression modules are inflated to securely grip the object, as \autoref{fig:control}(a.2). Next, to transport the object, as \autoref{fig:control}(a.3), the bottom compression module is deflated while the middle longitudinal module inflates, generating an axial displacement. Once the middle module has displaced the object, the bottom module is inflated again to re-grasp the object, and the top module is deflated. Finally, the middle module is deflated, the top module is re-inflated, and the process is repeated. This cyclical sequence of inflating and deflating the modules ensures that the object is progressively transported through the fundamental unit.

\subsection{Multi-Level Synchronization Through Pressure Feedback}

In the prior three--module case, the object was in contact with both compression modules, C--1 and C--3. However, as the system is scaled by stacking additional modules -- common in practical applications -- it becomes critical to implement a control strategy that adjusts actuation based on the object's position along the stacked modules. In the multi-level actuator station shown in \autoref{fig:control}(b.1), the base level consists of a fundamental working unit in \autoref{fig:control}(a.1). When additional longitudinal-compression modules (L--4, C--5, $\cdots$) are stacked on top of the base, higher levels are formed. The key challenge is coordinating the actuation of these upper levels to ensure smooth object transport. Since the object moves from bottom to top, the base level is actuated first, but determining when to activate the upper levels as the object approaches remains an outstanding question for optimizing actuation timing and efficiency.

To address this, we utilize pressure feedback from sensors embedded in the air chambers. When an object is grasped, the internal pressure within the chambers increases, generally at a slower rate compared to when no object is present. We exploit the pressure change rate -- signaling the presence of an object -- and monitor it in real-time to control the actuation sequence, outlined next, for object transport.

At $t=0$, the object to be transported is in touch with C--1 and C--3. In the closed-loop control, once C--3 inflates, we also inflate C--5 and monitor the pressure in C--5 during the inflation process. If the pressure change rate is significant different from the baseline, we infer that the object is in contact with C--5. As illustrated in \autoref{fig:control}(b.2), the system attempts to grasp a cylindrical object with radius $r_o = 0.7r$. The pressure change rate in C--5 without the object is 4.33 $\si{kPa/s}$, while with the object, the rate is 8.48 $\si{kPa/s}$, which is approximately twice the rate without an object.

Upon detecting consistent contact, the system activates the upper level, which consists of both the longitudinal and compression modules (L--4 and C--5), to work together with the compression module directly beneath the longitudinal module (C--3). This combined actuation forms another basic working unit for transporting the object. The process continues, triggering additional levels in sequence, until the object reaches the top. This method ensures smooth and synchronized object transport across the stacked modules.

For longer stacks, a similar approach will be employed to transition from L--4, C--5 to L--6, C--7, and so on.

\section{Results and Discussion}
\label{sec:Results}

\begin{figure*}[ht]
\centering
\includegraphics[width=\linewidth]{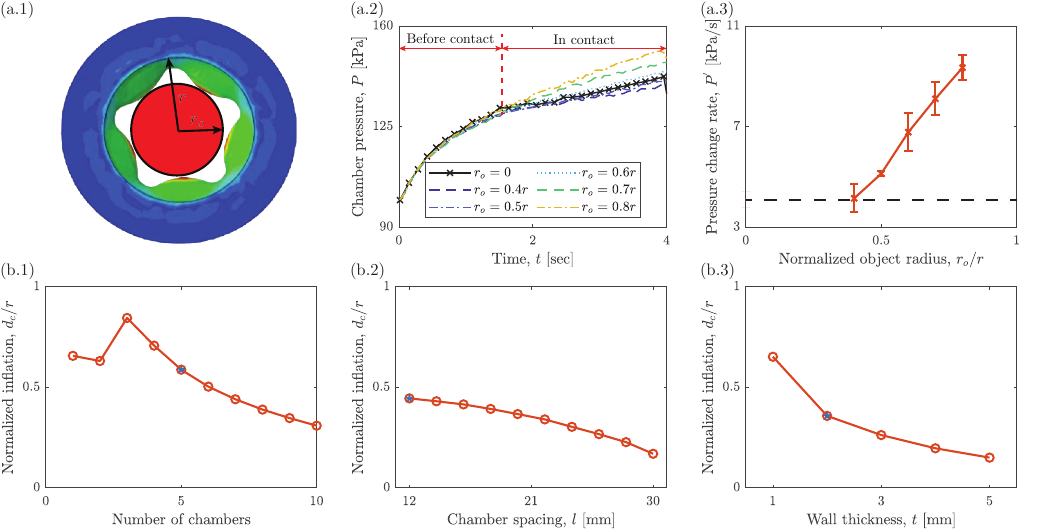}
\caption{Performance evaluation of the multi-stack actuator station. (a) Grasping capability characterization. (a.1) Schematic of object diameter versus inner radius. (a.2) Chamber pressure evolution when grasping objects of varying dimensions. (a.3) Pressure change rates during the in-contact stage. (b) Analysis of normalized inflation concerning the number of air chambers (b.1), chamber spacing (b.2), and wall thickness (b.3). The blue stars indicate the representative geometric parameters used in the experiments.}
\label{fig:result}
\end{figure*}

In this section, we delve into the performance of the multi-stack actuator station, focusing on how effectively the system can synchronize the actuation modules based on pressure feedback. The first part evaluates the capability of the system to detect the presence of cylindrical objects of varying diameters by observing pressure change rates during the grasping process. We assess when the system can reliably differentiate between cases with and without an object. Once this detection capability is established, we shift focus to optimizing the system’s grasping range by analyzing how the actuator's inflation displacement can be enhanced through geometrical design. A larger inflation displacement would enable the system to grasp thinner objects, thereby increasing the variety of objects that can be transported.

\subsection{Object Dimension and Pressure-Based Synchronization}

In exploring the dimensional limitations of objects that can be effectively transported, it is evident that the effective dimension of the object must not exceed the inner diameter of the actuation ring. Additionally, if the object's dimensions are too close to this diameter, inflation may be restricted, potentially leading to damage to the air chamber. Conversely, if the object is too small, the inflated compression module may not make adequate contact. In such cases, even if the system can grasp the object, the air chambers may inflate to the point of touching each other before making contact with the object, which hinders the generation of the necessary pressure differential for effective multi-level synchronization.

To determine the optimal object dimensions for effective grasp and transportation, as \autoref{fig:result}(a.1), we utilize cylindrical objects characterized by the radius $r_o$, ranging from 0.4 to 0.8 times the inner radius $r$ of the actuation ring. In \autoref{fig:result}(a.2), we observe an initial transition phase—termed the ``before contact stage" -- lasting approximately 1.5 seconds, during which the pressure curves exhibit similar trends regardless of the object's dimensions. Following this, in the contact stage, larger dimensions result in a higher pressure change rate due to increased squeezing.

To further characterize this phenomenon, we measure the pressure change rate, defined as $P' = dP/dt$, across varying cylinder radii, as shown in \autoref{fig:result}(a.3). The baseline is established with no object present, and we find that when $r_o = 0.4r$, $P'$ closely aligns with the baseline. This suggests that the system cannot reliably detect the presence of an object within the tubular tunnel, although the object can still be grasped and transported. Therefore, the effectiveness of multi-level synchronization through pressure feedback is constrained. Consequently, for reliable object transport, it is recommended to use objects with dimensions exceeding half the inner diameter of the ring.

\subsection{Optimizing Geometry for Maximum Inflation}
\label{sec:optimal_geometry}

From the previous subsection, we established the dimension range within which the actuation ring can effectively grasp and transport objects. A key factor in improving this performance is the radial inflation displacement -- since greater inflation increases the likelihood that the air chambers will contact the object, leading to an effective pressure change. Therefore, in this study, we analyze how the geometrical parameters of the actuation ring can be adjusted to maximize inflation displacement.

Finite element analysis is employed to explore this optimization due to its efficiency and low cost. We test various designs and evaluate their performance virtually, aiding in the refinement of the real system. To fully describe the geometry of the compression ring, we consider six key parameters, as illustrated in \autoref{fig:design}(a). While the inner radius $r$ and outer radius $R$ of the ring remain fixed, since they define the critical dimensions of the system, we explore how other parameters influence the inflation behavior, represented by normalized maximum inflation displacement $d_c/r$.

The step height $m$ defines the air tunnel dimension, affecting how efficiently air can inflate between the chambers. However, given the already small system dimensions, there is limited flexibility in adjusting this parameter without risking air flow impedance. As a result, we exclude this parameter from further optimization.

The key variables that characterize the air chamber distribution around the ring include the chamber length $l$ and chamber spacing $s$. Together with the number of air chambers $N$, the following relationship governs their arrangement:
\begin{equation*}
    (s + l) \times N = \pi(R+r).
\end{equation*}
Therefore, we investigate how variations in \( N \) and \( l \) influence inflation, with \( s \) coupled to these parameters while \( R \) and \( r \) remain fixed. 
As shown in \autoref{fig:result}(b.2), we observe that \( d_c/r \) generally decreases as \( N \) increases, primarily due to the increased air chamber volume, which causes the chambers to compress against each other. However, for \( N = 1 \) and \( N = 2 \), \( d_c/r \) is lower than for \( N = 3 \). This is because having too few chambers leads to non-uniform inflation, as previously discussed.
We conclude that an optimal \( N \) lies between 4 and 6, balancing inflation efficiency and uniformity.
Additionally, from \autoref{fig:result}(b.2), we observe that as the spacing \( l \) increases, \( d_c/r \) decreases. This is because chambers packed too densely inflate more slowly, leading to inefficient inflation-deflation cycles.

Besides, wall thickness also plays a crucial role in surface stiffness. Thinner walls inflate more easily, as \autoref{fig:result}(b.3), but are more prone to damage, such as sticking out or bursting under pressure. 

To balance the overall robustness and performance, we settle on an optimal design (marked as blue star) with $s = 28\si{mm}$, $l = 12\si{mm}$ (corresponding to $N = 5$), and $t = 2\si{mm}$. All experiments were conducted using this configuration.

\section{Conclusions}
\label{sec:Conclusions}

In conclusion, this study presents a bio-inspired pneumatic modular actuator designed for peristaltic transport, highlighting its potential for efficient object manipulation. By developing distinct actuation modules -- compression and longitudinal -- we have established a scalable, adaptable architecture that allows for seamless, cost-effective integration. The integration of pressure feedback mechanisms enhances the system’s ability to effectively detect and transport objects of varying sizes, optimizing its performance across different applications. 

Future work will focus on adapting the system for underwater environments, enabling its use in the collection of fragile specimens, such as coral reef. Additionally, efforts will be made to enhance its integration with existing robotic platforms, such as manipulators or Autonomous Underwater Vehicles (AUVs), to expand its applicability in various operational contexts.

\addtolength{\textheight}{-12cm}   % This command serves to balance the column lengths

%%%%%%%%%%%%%%%%%%%%%%%%%%%%%%%%%%%%%%%%%%%%%%%%%%%%%%%%%%%%%%%%%%%%%%%%%%%%%%%%

%%%%%%%%%%%%%%%%%%%%%%%%%%%%%%%%%%%%%%%%%%%%%%%%%%%%%%%%%%%%%%%%%%%%%%%%%%%%%%%%

%%%%%%%%%%%%%%%%%%%%%%%%%%%%%%%%%%%%%%%%%%%%%%%%%%%%%%%%%%%%%%%%%%%%%%%%%%%%%%%%

\section*{Acknowledgement}
We acknowledge financial support from the National Science Foundation under grant numbers 2047663 and 2332555.

%%%%%%%%%%%%%%%%%%%%%%%%%%%%%%%%%%%%%%%%%%%%%%%%%%%%%%%%%%%%%%%%%%%%%%%%%%%%%%%%
\bibliographystyle{ieeetr}
\bibliography{mybib}

\begin{thebibliography}{10}

\bibitem{patel2023physiology}
K.~S. Patel and A.~Thavamani, ``Physiology, peristalsis,'' in {\em StatPearls [Internet]}, StatPearls Publishing, 2023.

\bibitem{quillin1999kinematic}
K.~J. Quillin, ``Kinematic scaling of locomotion by hydrostatic animals: ontogeny of peristaltic crawling by the earthworm lumbricus terrestris,'' {\em Journal of Experimental Biology}, vol.~202, no.~6, pp.~661--674, 1999.

\bibitem{nakamura2008locomotion}
T.~Nakamura and T.~Iwanaga, ``Locomotion strategy for a peristaltic crawling robot in a 2-dimensional space,'' in {\em 2008 IEEE International Conference on Robotics and Automation}, pp.~238--243, IEEE, 2008.

\bibitem{tanaka2012mechanics}
Y.~Tanaka, K.~Ito, T.~Nakagaki, and R.~Kobayashi, ``Mechanics of peristaltic locomotion and role of anchoring,'' {\em Journal of the Royal Society Interface}, vol.~9, no.~67, pp.~222--233, 2012.

\bibitem{bursian2012structure}
A.~Bursian, ``Structure of autorhythmical activity of contractile systems,'' {\em Journal of Evolutionary Biochemistry and Physiology}, vol.~48, pp.~219--235, 2012.

\bibitem{brasseur1987fluid}
J.~G. Brasseur, ``A fluid mechanical perspective on esophageal bolus transport,'' {\em Dysphagia}, vol.~2, pp.~32--39, 1987.

\bibitem{gora2016tubular}
A.~G{\'o}ra, D.~Pliszka, S.~Mukherjee, and S.~Ramakrishna, ``Tubular tissues and organs of human body—challenges in regenerative medicine,'' {\em Journal of nanoscience and nanotechnology}, vol.~16, no.~1, pp.~19--39, 2016.

\bibitem{saga2004development}
N.~Saga and T.~Nakamura, ``Development of a peristaltic crawling robot using magnetic fluid on the basis of the locomotion mechanism of the earthworm,'' {\em Smart materials and structures}, vol.~13, no.~3, p.~566, 2004.

\bibitem{seok2012meshworm}
S.~Seok, C.~D. Onal, K.-J. Cho, R.~J. Wood, D.~Rus, and S.~Kim, ``Meshworm: a peristaltic soft robot with antagonistic nickel titanium coil actuators,'' {\em IEEE/ASME Transactions on mechatronics}, vol.~18, no.~5, pp.~1485--1497, 2012.

\bibitem{sensoy2021review}
I.~Sensoy, ``A review on the food digestion in the digestive tract and the used in vitro models,'' {\em Current research in food science}, vol.~4, pp.~308--319, 2021.

\bibitem{trivedi2008soft}
D.~Trivedi, C.~D. Rahn, W.~M. Kier, and I.~D. Walker, ``Soft robotics: Biological inspiration, state of the art, and future research,'' {\em Applied bionics and biomechanics}, vol.~5, no.~3, pp.~99--117, 2008.

\bibitem{rus2015design}
D.~Rus and M.~T. Tolley, ``Design, fabrication and control of soft robots,'' {\em Nature}, vol.~521, no.~7553, pp.~467--475, 2015.

\bibitem{mangan2002development}
E.~V. Mangan, D.~A. Kingsley, R.~D. Quinn, and H.~J. Chiel, ``Development of a peristaltic endoscope,'' in {\em Proceedings 2002 IEEE international conference on robotics and automation (cat. No. 02CH37292)}, vol.~1, pp.~347--352, IEEE, 2002.

\bibitem{menciassi2004sma}
A.~Menciassi, S.~Gorini, G.~Pernorio, and P.~Dario, ``A sma actuated artificial earthworm,'' in {\em IEEE International Conference on Robotics and Automation, 2004. Proceedings. ICRA'04. 2004}, vol.~4, pp.~3282--3287, IEEE, 2004.

\bibitem{nakamura2006development}
T.~Nakamura, ``Development of a peristaltic crawling robot using servomotors based on the locomotion mechanism of earthworms,'' in {\em Proc. Of IEEE International Conference on Robotics and Automation (ICRA), 2006}, pp.~4342--4344, 2006.

\bibitem{nakamura2006development1}
T.~Nakamura, T.~Kato, T.~Iwanaga, and Y.~Muranaka, ``Development of a peristaltic crawling robot based on earthworm locomotion,'' {\em Journal of Robotics and Mechatronics}, vol.~18, no.~3, pp.~299--304, 2006.

\bibitem{saga2004prototype}
N.~Saga, ``A prototype of peristaltic robot using pneumatic artificial muscle,'' {\em Intelligent autonomous system}, vol.~8, no.~1, pp.~85--95, 2004.

\bibitem{kim2005earthworm}
B.~Kim, S.~Park, C.~Y. Jee, and S.-J. Yoon, ``An earthworm-like locomotive mechanism for capsule endoscopes,'' in {\em 2005 IEEE/RSJ international conference on intelligent robots and systems}, pp.~2997--3002, IEEE, 2005.

\bibitem{peng2024peristaltic}
Y.~Peng, H.~Nabae, Y.~Funabora, and K.~Suzumori, ``Peristaltic transporting device inspired by large intestine structure,'' {\em Sensors and Actuators A: Physical}, vol.~365, p.~114840, 2024.

\bibitem{dirven2013design}
S.~Dirven, F.~Chen, W.~Xu, J.~E. Bronlund, J.~Allen, and L.~K. Cheng, ``Design and characterization of a peristaltic actuator inspired by esophageal swallowing,'' {\em IEEE/ASME Transactions on Mechatronics}, vol.~19, no.~4, pp.~1234--1242, 2013.

\bibitem{hashem2020design}
R.~Hashem, M.~Stommel, L.~K. Cheng, and W.~Xu, ``Design and characterization of a bellows-driven soft pneumatic actuator,'' {\em IEEE/ASME Transactions on Mechatronics}, vol.~26, no.~5, pp.~2327--2338, 2020.

\bibitem{dang2019soft}
Y.~Dang, M.~Stommel, L.~K. Cheng, and W.~Xu, ``A soft ring-shaped actuator for radial contracting deformation: Design and modeling,'' {\em Soft robotics}, vol.~6, no.~4, pp.~444--454, 2019.

\bibitem{kim2011epidermal}
D.-H. Kim, N.~Lu, R.~Ma, Y.-S. Kim, R.-H. Kim, S.~Wang, J.~Wu, S.~M. Won, H.~Tao, A.~Islam, {\em et~al.}, ``Epidermal electronics,'' {\em science}, vol.~333, no.~6044, pp.~838--843, 2011.

\end{thebibliography}

\end{document}